\documentclass[11pt,a4paper]{article}
\usepackage[hyperref]{acl2020}
\usepackage{times}
\usepackage{booktabs}
\usepackage{latexsym}
\usepackage{float}
\usepackage{graphicx}
\usepackage{subcaption}
\usepackage{amsmath}
\usepackage{multirow}

\usepackage{microtype}

\aclfinalcopy 

\title{WLV-RIT at GermEval 2021: Multitask Learning with Transformers to Detect Toxic, Engaging, and Fact-Claiming Comments}

\author{Skye Morgan\textsuperscript{1}, Tharindu Ranasinghe\textsuperscript{2}, Marcos Zampieri\textsuperscript{1} \\
  \textsuperscript{1}Rochester Institute of Technology, USA\\
  \textsuperscript{2}University of Wolverhampton, UK \\
  \texttt{sdm9815@rit.edu} 
  }

\date{}

\begin{document}
\maketitle
\begin{abstract}
This paper addresses the identification of toxic, engaging, and fact-claiming comments on social media. We used the dataset made available by the organizers of the GermEval-2021 shared task containing over 3,000 manually annotated Facebook comments in German. Considering the relatedness of the three tasks, we approached the problem using large pre-trained transformer models and multitask learning. Our results indicate that multitask learning achieves performance superior to the more common single task learning approach in all three tasks. We submit our best systems to GermEval-2021 under the team name WLV-RIT. 
\end{abstract}

\section{Introduction}
\label{sec:intro}

The popularity and accessibility associated with social media have greatly promoted user-generated content. At the same time, social media sites have increasingly become more prone to offensive content~\cite{hada-etal-2021-ruddit,zhu-bhat-2021-generate,bucur-etal-2021-exploratory}. As such, identifying the toxic language in social media is a topic that has gained, and continues to gain traction. Research surrounding the problem of offensive content has centered around the application of computational models that can identify various forms of negative content such as hate speech ~\cite{malmasi2018challenges,nozza-2021-exposing}, abuse \cite{corazza-etal-2020-hybrid}, aggression ~\cite{kumar2018benchmarking,kumar-etal-2020-evaluating}, and cyber-bullying ~\cite{rosa2019automatic,cheng-etal-2021-mitigating,salawu-etal-2021-large}.

GermEval-2021~\cite{germeval2021overview} focuses on identifying multiple types of comments in social media. This year's shared task is divided into three distinct classifications of comments:  {\it i}) Toxic, {\it ii}) Engaging, and {\it iii}) Fact-Claiming. Like previous GermEval shared tasks \cite{germeval-2019}, the detection of toxic content remains an integral part of GermEval-2021. Regarding engaging comments, there is an increasing desire from community managers as well as moderators to identify valuable user content \cite{kolhatkar2017using,napoles2017finding}. More particularly, rational comments that serve to encourage readers to engage in a discussion. In a similar light, identifying fact-claiming comments is equally important as platforms need to consistently review and verify user-generated content to uphold their responsibility as information distributors \cite{mihaylova2018fact,shaar2020overview}. 

We pose that multitask learning (MTL) is a suitable approach for this year's GermEval as it enables what is learned from each task to aid in the learning of other tasks. The current state-of-the-art approach for offensive language identification is neural transformers modeled using single task learning (SLT) \cite{liu-etal-2019-nuli,ranasinghe-etal-2020-multilingual}. It is well-known that training large neural transformer models often result in long processing times. As GermEval-2021 features three related tasks, from a performance standpoint, we pose that training a model jointly on three tasks is likely to be computationally more efficient than training three models in isolation. Moreover, as GermEval-2021 provides a single dataset for the three tasks, MTL can also be used to help improving performance across tasks. As such, we introduce multitask learning whereby one model can predict all three tasks as an alternative approach.

In this paper, we present the methods and results of the WLV-RIT submission to the GermEval-2021 shared task. We explore transformer architectures in two different environments, single task learning and multitask learning, and describe them in detail in Section \ref{sec:methods}. We perform several experiments using three transformer models that support German and evaluate their performance on the GermEval-2021 dataset.

\section{Related Work}
\label{sec:RW}

The identification of offensive language in online discussions is an extensive topic that has become popular over the past several years. The majority of the research related to this topic is centered on English data due to the availability of annotated datasets \cite{OLID,rosenthal2020}. Notwithstanding this, offensive language datasets are being annotated in other languages. Researchers have examined offensive content across multiple social media platforms and have both annotated and utilized data from different languages such as Greek ~\cite{pitenis2020}, Marathi~\cite{mold}, Italian ~\cite{chiril-etal-2019-multilingual}, Portuguese ~\cite{fortuna2019hierarchically,vargas2021contextual}, Arabic ~\cite{mubarak2020arabic}, Turkish ~\cite{coltekin2020}, and multiple languages of India~\cite{ranasinghe2021MDPI}. 

Past approaches to tackling the problem of offensive content on social media have relied on using a variety of computational models ranging from traditional machine learning classifiers such as Logistic Regression and SVMs ~\cite{malmasi2018challenges}, to various deep learning models ~\cite{de-gibert-etal-2018-hate}. SemEval-2019 Task 5 (HatEval) ~\cite{basile2019semeval} presented the challenge of detecting the presence of hate speech and identifying further features in hateful contents, which included two sub-tasks. For subtask A, which was the hate speech (HS) category, the best performance was achieved by training a support vector machine (SVM) model with a radial basis function (RBF) kernel. Several other high scoring teams used a convolutional neural network (CNN) which was traditionally the most popular approach to this topic ~\cite{hettiarachchi-ranasinghe-2019-emoji}. For TRAC-1 ~\cite{kumar2018benchmarking}, the challenge was to develop a classifier that could discriminate between three levels of aggression in social media. The results showed that with careful consideration, classifiers like SVM and even random forest could perform at par with deep neural networks. However, in the end, more than half of the top 15 systems were trained on neural networks which demonstrates the approach's effectiveness.

\begin{table*}
\centering
\scalebox{0.92}{
\begin{tabular}{p{13cm}cccc@{}}
\toprule
\multicolumn{1}{c}{\textbf{Comment}} &
  \multicolumn{1}{c}{\textbf{Sub1}} &
  \multicolumn{1}{c}{\textbf{Sub2}} &
  \multicolumn{1}{c}{\textbf{Sub3}} \\ \midrule
"Die AfD sind genau so neoliberal und kapitalistische Zerstörer unserer Heimat, wie die CDU, CSU, FDP, SPD und Grüne auch." & 1 & 0 & 0 \\
"Sarazin ist ein rechtsradikaler Mensch. Ein Menschenhasser. Sie kennen nur Zerstörung. Die Geschichte hat es gezeigt."     & 1 & 0 & 1 \\
"@USER, du hast das Thema im Kern nicht verstanden"                                                                         & 0 & 0 & 1 \\
"Ich frage dich, verlassen Menschen gerne ihre Heimat?"                                                                     & 0 & 0 & 0 \\ \bottomrule
\end{tabular}}
\caption{Annotation examples of four different Facebook user comments. Sub1 represents toxic comments, Sub2 stands for engaging comments, and Sub3 stands for fact claiming.}
\label{tab:Annotation Examples}
\end{table*} 

The introduction of  BERT ~\cite{devlin2019bert} spurred the use of pre-trained transformer models for classifying offensive speech~\cite{ranasinghemudes}. As a result, neural transformer based language models have increasingly become more popular in offensive language identification. The use of pre-trained BERT models, as well as BERT-based models, was shown to be able to achieve competitive performance in popular competitions such as OffensEval ~\cite{offenseval,zampieri-etal-2020-semeval}. Language-specific and multilingual models have also been introduced to assist NLP research in various languages such as GBERT for German ~\cite{chan-etal-2020-germans}, AraBERT for Arabic ~\cite{antoun-etal-2020-arabert}, and the multilingual XLM-R ~\cite{conneau2019unsupervised} that has been been applied to offensive language identification \cite{ranasinghe-etal-2020-multilingual,ranasingheTALLIP}.

\section{Data}
\label{sec:data}

In the GermEval-2021 dataset, the focus has been extended beyond the identification of offensive comments to include two additional classes: engaging comments that can motivate readers to participate in conversations, and fact-claiming comments. The dataset for this iteration of GermEval comprises over 3,000 Facebook user comments that have been extracted from the page of a political talk show of a German television broadcaster. The training dataset has a total of 3,244 instances and comprises 1,074 instances without any toxic, engaging or fact claiming content. In Table 1, we present four different Facebook user comments along with their annotation.

\begin{table}[H]
\centering
\scalebox{0.92}{
\begin{tabular}{@{}cccccc@{}}
\toprule
\multicolumn{1}{l}{\textbf{Toxic}} & \multicolumn{1}{l}{\textbf{Engaging}} & \multicolumn{1}{l}{\textbf{Fact-Claiming}} & \multicolumn{1}{l}{\textbf{Training}} \\ \midrule
0   & 0 & 0 & 1074 \\
1   & 0 & 0 & 739  \\
0   & 1 & 0 & 239  \\
1   & 1 & 0 & 89   \\
0   & 1 & 1 & 403  \\
1   & 0 & 1 & 160  \\
0   & 0 & 1 & 406  \\
1   & 1 & 1 & 134  \\ \hline
All &   &   & 3244 \\ \bottomrule
\end{tabular}
}
\caption{GermEval 2021 - Training Set User Comment Distribution}
\label{tab:stats}
\end{table}

\renewcommand{\arraystretch}{1.2}
\begin{table}[!ht]
\centering
\scalebox{0.92}{
\begin{tabular}{ll}
\toprule
\bf Parameter &   \bf Value \\ \hline
learning rate$^{\ddag}$ & $1e^{-5}$\\
number of epochs$^{\ddag}$ & $3$\\
adam epsilon & $1e^{-8}$\\
warmup ratio & 0.1\\
warmup steps & 0\\
max grad norm & 1.0\\
max seq. length & 120\\
gradient accumulation steps & 1\\
\bottomrule
\end{tabular}
}
\caption{Hyperparameter specifications. The optimised hyperparameters are marked with $\ddag$ and their optimal values are reported. The rest of the hyperparameter values are kept as constants.}
\label{tab:params}
\end{table}

\begin{table*}
\centering
\scalebox{0.88}{
\begin{tabular}{l|c|ccc|ccc|ccc}

\hline
                                     & & \multicolumn{3}{c|}{\textbf{Toxic}} & \multicolumn{3}{c|}{\textbf{Engaging}}             & \multicolumn{3}{c}{\textbf{Fact-Claiming}}            \\ \hline
\multicolumn{1}{l|}{\textbf{Model}} & \textbf{Environment} & \textbf{P}   & \textbf{R}   & \textbf{F1}   & \textbf{P} & \textbf{R} & \textbf{F1}               & \textbf{P} & \textbf{R} & \textbf{F1}      \\ \hline
\multirow{ 4}{*}{\textit{mBERT}}& STL & 0.4897         & 0.4421         & 0.4500          & 0.5421       & 0.5310       & 0.5380 & 0.5532       & 0.5093       & 0.5511    \\
& LM + STL & 0.4921         & 0.4432         & 0.4512          & 0.5436       & 0.5314       & 0.5398 & 0.5669       & 0.5101       & 0.5521    \\
 & MTL & 0.5042         & 0.4449         & 0.4551          & 0.5472       & 0.5325       & 0.5401 & 0.5702       & 0.5113       & 0.5532    \\
 & LM + MTL & 0.5063         & 0.4543         & 0.4665          & 0.5542       & 0.5341       & 0.5442 & 0.5732       & 0.5231       & 0.5555    \\
 \hline

\multirow{ 4}{*}{\textit{gBERT}}  & STL & 0.6449         & 0.5801         & 0.6102          & 0.6449       & 0.6312       & 0.6342 & 0.6812       & 0.6752       & 0.6852    \\
& LM + STL & 0.6552         & 0.5841         & 0.6173          & 0.6254       & 0.6442       & 0.6354 & 0.6821       & 0.6779       & 0.6872    \\
 & MTL & 0.7001         & 0.6321         & 0.6654          & 0.6777       & 0.6931       & 0.6841  & 0.7311       & 0.7211       & 0.7352    \\
 & LM + MTL$^{\ddagger}$ & 0.7124         & 0.6456         & 0.6796          & 0.6827       & 0.7027       & 0.6926 & 0.7450       & 0.7495       & \textbf{0.7472}    \\
 \hline
\multirow{ 4}{*}{\textit{gELECTRA}}   & STL & 0.6551         & 0.5991         & 0.6227          & 0.6391       & 0.6482       & 0.6431 & 0.6954       & 0.7002       & 0.7045    \\
& LM + STL & 0.6651         & 0.6078         & 0.6321          & 0.6422       & 0.6561       & 0.6555 & 0.7021       & 0.7102       & 0.7100    \\
 & MTL$^{\ddagger}$ & 0.7256         & 0.6603         & 0.6914          & 0.6895       & 0.6999       & \textbf{0.6947}  & 0.7530       & 0.7407       & 0.7468    \\
 & LM + MTL$^{\ddagger}$ & 0.7542  & 0.6732         & \textbf{0.7112}          & 0.6944       & 0.6924       & 0.6934 & 0.7354       & 0.7383       & 0.7369    \\

\hline
\end{tabular}
}
\caption{Results for the evaluation set in each task with Transformer models. For each model, Precision (P), Recall (R), and F1 are reported on all tasks. The best result for each task has been marked with bold considering F1. The experiments we submitted are marked with ${\ddagger}$}
\label{table:dev_result}
\end{table*}

\section{Methods}
\label{sec:methods}

Considering the success that neural transformers have demonstrated across various natural language processing tasks ~\cite{uyangodage-etal-2021,jauhiainen-etal-2021-comparing,hettiarachchi-ranasinghe-2020-brums} including offensive language identification ~\cite{ranasinghe-etal-2020-multilingual,ranasinghemudes,dai-etal-2020-kungfupanda} we used transformers to tackle this task too. We explored transformer architectures in two different environments; single task learning and multi task learning. 

\paragraph{Single Task Learning (STL)} For the STL environment we trained three classification models based on transformers. By utilizing the hidden representation of the classification token (\textsc{CLS}) in the transformer model, we predict the target labels (toxic/non-toxic, engaging/non-engaging, fact-claiming, non-fact-claiming) by applying a linear transformation followed by the softmax activation ($\sigma$):
\begin{align}
    \hat{\mathbf{y}}_{task} = \sigma(\mathbf{W}_{[CLS]} \cdot \mathbf{h}_{[CLS]} + \mathbf{b}_{[CLS]})
\end{align}
where $\cdot$ denotes matrix multiplication, $\mathbf{W}_{[CLS]} \in \mathcal{R}^{D \times 3}$, $\mathbf{b}_{[CLS]} \in \mathcal{R}^{1 \times 2}$, and $D$ is the dimension of the input activation layer $\mathbf{h}$. $\hat{\mathbf{y}}_{task}$ is the predicted value of any of the three tasks. 

\begin{figure}[!ht]
\centering
  \begin{subfigure}[b]{6cm}
    \centering\includegraphics[width=5.1cm]{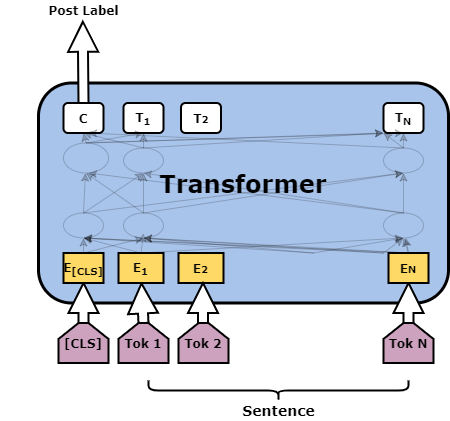}
    \caption{STL Architecture}
    \label{fig:stl}
  \end{subfigure}
    \begin{subfigure}[b]{6cm}
    \centering\includegraphics[width=5.1cm]{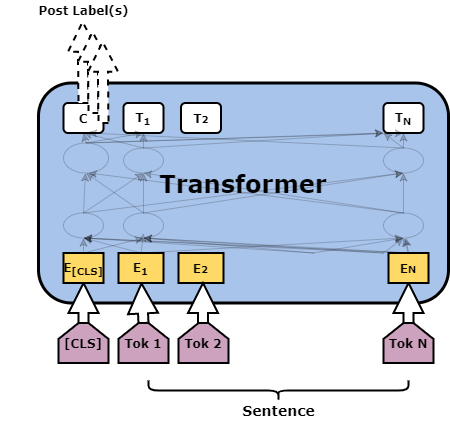}
    \caption{MTL Architecture}
    \label{fig:mtl}
  \end{subfigure}
    
\caption{The STL (top) and MTL (bottom) transformer-based architectures experimented with the GermEval-2021 dataset.}
\label{fig:architecture}
\end{figure}

We construct three separate classification models minimising the cross-entropy loss for each of the three tasks as defined in the Equation \ref{eqn:stl_loss}, where $y_{toxic}$, $y_{engage}$  and $y_{fact}$ represent ground truth labels of each task. These particular losses are:
\begin{align}
    \mathcal{L}_{toxic} &=-\sum^2_{i=1} \Big( \mathbf{y}_{toxic} \otimes \log( \hat{\mathbf{y}}_{toxic} ) \Big)[i] \nonumber \\
      \mathcal{L}_{engage} &=-\sum^2_{i=1} \Big( \mathbf{y}_{engage} \otimes \log( \hat{\mathbf{y}}_{engage} ) \Big)[i] \nonumber \\
       \mathcal{L}_{fact} &=-\sum^2_{i=1} \Big( \mathbf{y}_{fact} \otimes \log( \hat{\mathbf{y}}_{fact} ) \Big)[i]
\label{eqn:stl_loss}
\end{align}
where $\mathbf{v}[i]$ retrieves the $i$th item in a vector $\mathbf{v}$ and $\otimes$ indicates element-wise multiplication. The corresponding STL architecture is shown in Figure \ref{fig:stl}.

\paragraph{Multi Task Learning (MTL)} MTL was introduced as an approach to inductive transfer ~\cite{caruana_multitask_1997}. The main goal of which was to improve generalization performance on a current task after having learned a different but related concept on a previous task.  MTL is quite efficient as one model can be utilized to predict multiple tasks so long as they are related. In hate speech and offensive language detection, MTL has been shown to outperform single-task environments as well as learn task efficiently with the presence of little labelled data per-task ~\cite{djandji-etal-2020-multi}. Despite this, MTL has not been used much in the context of offensive language detection. As such, we decided to use multitask learning to compare the performance within the two different environments using different transformer models. We used the transformer as the base model for our MTL approach. Our approach will learn the three tasks jointly, i.e., Toxic comment detection, Engaging comment detection and Fact-claiming comment detection. The implemented architecture shares the hidden layers between the tasks. The shared portion includes a transformer model that learns shared information across the tasks by minimizing a combined loss. We assign equal importance to each task in our experiments. The full loss is: 

\begin{equation}
\label{eqn:mad_loss_germeval}
   \mathcal{L}_{multi} = \frac{\mathcal{L}_{toxic} + \mathcal{L}_{engage} + \mathcal{L}_{fact}}{3} \mbox{.}
\end{equation}

\noindent The task-specific classifiers receive input from the last hidden layer of the transformer language model and predict the output for the tasks. The corresponding MTL architecture is shown in Figure \ref{fig:mtl}

\begin{table*}
\centering
\scalebox{0.88}{
\begin{tabular}{l|c|ccc|ccc|ccc}

\hline
                                     & & \multicolumn{3}{c|}{\textbf{Toxic}} & \multicolumn{3}{c|}{\textbf{Engaging}}             & \multicolumn{3}{c}{\textbf{Fact-Claiming}}            \\ \hline
\multicolumn{1}{l|}{\textbf{Model}} & \textbf{Environment} & \textbf{P}   & \textbf{R}   & \textbf{F1}   & \textbf{P} & \textbf{R} & \textbf{F1}               & \textbf{P} & \textbf{R} & \textbf{F1}      \\ \hline
\multirow{ 4}{*}{\textit{mBERT}}& STL & 0.5081         & 0.4672         & 0.4781          & 0.5689       & 0.5561       & 0.5555 & 0.5763       & 0.5286       & 0.5761    \\
& LM + STL & 0.5162         & 0.4657         & 0.4782          & 0.5698       & 0.5561       & 0.5568 & 0.5871       & 0.5389       & 0.5780    \\
 & MTL & 0.5284         & 0.4672         & 0.4781          & 0.5690       & 0.5571       & 0.5678 & 0.5901       & 0.5364       & 0.5782    \\
 & LM + MTL & 0.5243         & 0.4763         & 0.4871          & 0.5762       & 0.5590       & 0.5601 & 0.5983       & 0.5482       & 0.5782    \\
 \hline

\multirow{ 4}{*}{\textit{gBERT}}  & STL & 0.6692         & 0.6092         & 0.6354          & 0.6678       & 0.6572       & 0.6532 & 0.7095       & 0.6982       & 0.7011    \\
& LM + STL & 0.6752         & 0.6072         & 0.6342          & 0.6453       & 0.6683       & 0.6572 & 0.7063       & 0.6982       & 0.7041    \\
 & MTL & 0.7223         & 0.6532         & 0.6842          & 0.6954       & 0.7132       & 0.7041  & 0.7553       & 0.7493       & 0.7562    \\
 & LM + MTL$^{\ddagger}$ & 0.7321         & 0.6654         & 0.6941          & 0.7041       & 0.7298       & 0.7145 & 0.7653       & 0.7602       & 0.7652    \\
 \hline
\multirow{ 4}{*}{\textit{gELECTRA}}   & STL & 0.6752         & 0.6111         & 0.6498          & 0.6531       & 0.6679       & 0.6609 & 0.7178       & 0.7285       & 0.7265    \\
& LM + STL & 0.6874         & 0.6231         & 0.6562          & 0.6666       & 0.6742       & 0.6731 & 0.7231       & 0.7303       & 0.7367    \\
 & MTL$^{\ddagger}$ & 0.7456         & 0.6802         & 0.7132          & 0.7001       & 0.7101       & \textbf{0.7198}  & 0.7754       & 0.7652       & \textbf{0.7653}    \\
 & LM + MTL$^{\ddagger}$ & 0.7853  & 0.6997         & \textbf{0.7342}          & 0.7132       & 0.7156  & 0.7190 & 0.7542       & 0.7563       & 0.7590    \\

\hline
\end{tabular}
}
\caption{Results for the test set in each task with Transformer models. For each model, Precision (P), Recall (R), and F1 are reported on all tasks. The best result for each task has been marked with bold considering F1. The experiments we submitted are marked with ${\ddagger}$}
\label{table:test_result}
\end{table*}

\section{Experimental Setup}
\label{sec:Experimental_setup}
We performed experiments using three transformer models that support German; mBERT \cite{devlin2019bert}, German BERT-large (gBERT) \cite{chan-etal-2020-germans} and German Electra-large (gELECTRA) \cite{chan-etal-2020-germans} transformer models available in the HuggingFace model repository \cite{wolf-etal-2020-transformers}. 

We used an Nvidia Tesla K80 GPU to train the models. We divided the input dataset into a training set and a validation set using 0.8:0.2 split. We predominantly fine-tuned the learning rate and the number of epochs of the classification model manually to obtain the best results for the validation set. We obtained $1e^-5$ as the best value for the learning rate and 3 as the best value for the number of epochs. We used a batch size of 8 for the training process and the model was evaluated after every 100 batches. We performed \textit{early stopping} if the validation loss did not improve over 10 evaluation steps. The rest of the hyperparameters which we kept as constants are mentioned in the Table \ref{tab:params}. For both STL and MTL we finetuned the considered transformer model on the GermEval 2021 training set using Masked Language Modeling (MLM) \cite{devlin2019bert} objective which we call as Language Modeling (LM). When performing training, we trained five models with different random seeds and considered the majority-class self ensemble mentioned in \citet{hettiarachchi-ranasinghe-2020-infominer} to get the final predictions.

\section{Results}
\label{sec:results}

We show the results for the evaluation set in Table \ref{table:dev_result}. In all the experimented transformer models, the MTL approach outperformed the STL approach. Furthermore in most scenarios, the systems that included a LM component outperformed those without the LM component. This corroborates the findings of previous research in offensive language identification \cite{ranasinghe2019brums}. gBERT and gELECTRA models clearly outperformed mBERT in all the tasks. For the Task 1, gELECTRA model with LM and MTL achieved the best result with 0.7342 F1 score, for the Task 2 gELECTRA model with MTL, without LM achieved the best result with 0.7198 F1 score and for the Task 3 too, the same model achieved the best result with 0.7653 F1 score. Considering the overall performance we selected three best models for the submission; gELECTRA with LM+MTL, gELECTRA with MTL and gBERT with LM+MTL. 

The official leaderboard of the competition was not yet released at the time of writing this paper, therefore, after the organizers released the gold labels for the test set, we calculated the Precision, Recall, and F1 values for the test set. The results are shown in Table \ref{table:test_result}. As shown in the results, the three models we selected provided the top three results for the test set too. MTL consistently outperformed STL in all the tasks with all the transformer models we experimented.

\section{Conclusion and Future Work}
\label{sec:conclusion}
In this paper, we presented the WLV-RIT entry to GermEval-2021. GermEval-2021 provided participants with the opportunity of testing computational models to identify toxic, engaging, and fact claiming comments. We experimented with neural transformer models in STL environment and MTL environment. MTL environment consistently outperformed STL suggesting that the use of shared learning methods improves the performance of individual tasks. Furthermore, we observed that pre-trained language-specific transformer models trained for German such as gBERT and gElectra outperform mBERT. Finally, in addition to the transformer-based MTL approach, we could observe that the use of language modelling led performance improvement in some of the tasks.  

In the future, we would like to carry out an error analysis on the output of our systems to better understand the impact and limitations of MTL for these three tasks. Finally, we would like to experiment with multi-task learning in other languages, particularly low-resource languages for which only limited language resources are available.

\section*{Acknowledgments}

The authors would like to thank the GermEval-2021 organizers for organizing this interesting shared task and for making the dataset available. 

\bibliography{custom}
\bibliographystyle{acl_natbib}

\end{document}